\documentclass[acmtog]{acmart}

\usepackage{booktabs} 

\usepackage[ruled]{algorithm2e} 

\SetAlFnt{\small}
\SetAlCapFnt{\small}
\SetAlCapNameFnt{\small}
\SetAlCapHSkip{0pt}

\acmJournal{TOG}

\copyrightyear{2026}
\acmYear{2026}
\setcopyright{cc}
\setcctype{by}
\acmConference[SA Conference Papers '26]{SIGGRAPH Asia 2026 Conference Papers}{December 01--04, 2026}{Kuala Lumpur, Malaysia}
\acmBooktitle{SIGGRAPH Asia 2026 Conference Papers (SA Conference Papers '26), December 01--04, 2026, Kuala Lumpur, Malaysia}
\acmDOI{10.1145/3829340.3842168}
\acmISBN{979-8-4007-2842-6/2026/12}

\begin{document}
\title[Smart-Insertion-V: Video Insertion via a Dual-Stream Framework]{Smart-Insertion-V: Photorealistic Video Insertion via a Closed-Loop Feedback Dual-Stream Framework}

\author{Xiao Cao}
\orcid{0000-0001-9704-9633}
\email{xiaocao@u.nus.edu}
\affiliation{%
 \institution{National University of Singapore}
 \country{Singapore}}

 \author{Yansong Qu}
\orcid{0009-0003-4325-6858}
\affiliation{%
 \institution{Xiamen University}
 \country{China}}

 \author{Xiangzhen Chang}
\orcid{0009-0009-7794-8602}
\affiliation{%
 \institution{TeleAI}
 \country{China}}

 \author{Wen Xiao}
\orcid{0009-0006-8767-0702}
\affiliation{%
 \institution{National University of Singapore}
 \country{Singapore}}

 \author{Heyuan Li}
\orcid{0000-0001-8101-2916}
\affiliation{%
 \institution{Chinese University of Hong Kong, Shenzhen}
 \country{China}}

  \author{Jiakui Hu}
\orcid{0009-0000-7740-6373}
\affiliation{%
 \institution{Peiking University}
 \country{China}}

  \author{Liying Yang}
\orcid{0000-0001-6347-7736}
\affiliation{%
 \institution{Macau University of Science and Technology}
 \country{China}}

  \author{Jialun Liu}
  \email{liujialun95@gmail.com}
\orcid{0000-0002-6879-7648}
\affiliation{%
 \institution{TeleAI}
 \country{China}}
 \authornote{Corresponding author.}
 \authornote{Project leader.}

  \author{Zhiyong Huang}
  \email{dcshuang@nus.edu.sg}
\orcid{0000-0002-1931-7775}
\affiliation{%
 \institution{National University of Singapore}
 \country{Singapore}}
 \authornotemark[1]

\author{Xuelong Li}
\orcid{0000-0002-0019-4197}
\affiliation{%
 \institution{TeleAI}
 \country{China}}



\begin{abstract}
Mask-free video object insertion has emerged as a challenging task, requiring harmonious integration of reference objects into source videos. However, existing methods struggle when references exhibit severe stylistic domain gaps with the source scene. 
%
To overcome this, we propose \textit{\textbf{Smart-Insertion-V}}, an end-to-end \textbf{Dual-Stream} framework that concurrently conducts video insertion and image style transfer. 
Within this framework, the image stream synchronously guides the video generation process, while a \textbf{Closed-loop Feedback} mechanism is further incorporated to ensure robust insertion. Inevitably, integrating these diverse conditioning signals results in feature entanglement and style leakage. 
To tackle this issue, we design \textbf{Dual-World-View RoPE} to distinguish different signals via spatial-temporal offsets without incurring heavy training overhead. 
Furthermore, to facilitate spatial grounding and stylistic adaptation, we introduce a \textbf{Decoupled Guidance Module} that leverages a Vision-Language Model for semantic reasoning while preserving original temporal guidance with native text encoder. 
%
To bridge data gap for harmonious reference insertion task, we propose a data curation pipeline and will release an \textbf{open-source dataset}.
Experiments demonstrate that our method can insert objects into plausible positions while achieving the most harmonious results. Code and dataset will be released upon acceptance.
\end{abstract}

\begin{teaserfigure}
    \centering
    \includegraphics[width=\textwidth]{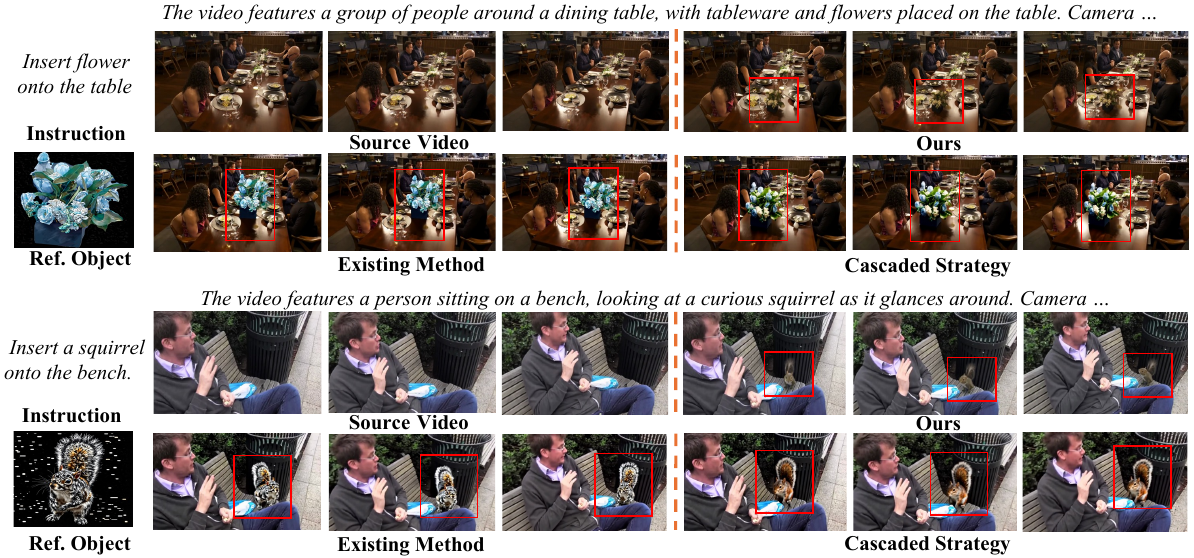}
    \caption{Our task aims to insert a raw reference object into a video. Existing image-based video insertion models and cascaded strategies (i.e., first performing image editing and then applying video insertion) fail to achieve satisfactory visual harmonization, while proposed Smart-Insertion-V inserts object into a plausible position with harmonized effect.}
    \label{fig:teaser}
\end{teaserfigure}

%
%

%
%


\maketitle

\section{Introduction}
\label{sec:intro}
Recent advances in diffusion-based video generation~\cite{diffusion1,diffusion2,diff1,diff2,diff3,diff4,diff5,diff6} have enabled increasingly powerful video editing capabilities~\cite{dragvideo,videoedit1,videoedit2,videoedit3,videoedit4,edit1,edit2,edit3,edit4,edit5,edit6,edit7}. Among them, video object insertion-seamlessly integrating a reference object into a source video-remains a challenging problem. Beyond simply placing an object into a scene, successful mask-free insertion requires jointly reasoning about spatial plausibility, adapting object appearance to match the scene style, and maintaining temporal consistency. This challenge is further amplified when the reference image is visually incompatible with the source video.

Existing methods can be broadly categorized into three groups: explicit spatial guidance methods, mask-free methods, and cascaded methods. They remain limited in spatial reasoning, reference-scene harmonization, or robustness across multiple stages.
Explicit spatial guidance methods~\cite{getinvideo,videoanydoor,anything,insert1,insert2,ma2024followpose,ma2025controllable} rely on user-specified priors, such as masks or control points, to localize the insertion region. Such dependence introduces additional annotation burden and limits practical applicability, while models' insufficient scene understanding still leads to implausible object attributes, including inconsistent scale, perspective, or placement.
Mask-free methods~\cite{omniinsert,univideo,teleomni} successfully eliminate manual spatial annotations. However, raw references frequently exhibit severe domain gaps with the target video, including discrepancies in lighting and style. By generating videos conditioned on unprocessed reference images without imposing explicit constraints, these methods often suffer from contextual disharmony between the inserted object and the source video.
A straightforward remedy is to employ a cascaded pipeline, which first adapts the reference image to the scene via an image editing model before performing video insertion. However, this decoupled approach is inherently prone to error accumulation. Artifacts or inaccurate adaptations introduced during the initial editing phase are propagated and amplified throughout the video generation process, severely degrading both visual quality and temporal coherence.

In this work, we propose Smart-Insertion-V, an end-to-end mask-free video insertion framework that seamlessly inserts raw reference objects into target videos under the guidance of coarse text prompts.
To address spatial perception challenges, we introduce a Decoupled Guidance Module (DGM). Specifically, a Vision-Language Model (VLM)~\cite{lee_2026tokag,teleomni,univideo} is incorporated to comprehend the scene's spatial layout and resolve style discrepancies. Meanwhile, the native T5 text encoder is retained as a motion reasoner to preserve temporal dynamics.
Furthermore, we propose an innovative dual-stream architecture consisting of a video stream and an image stream. While the video stream focuses on the core video insertion, the image stream concurrently performs reference image style transfer. Our core insight is that the image harmonization process can synchronously guide the video generation. Specifically, the image stream provides intermediate features that capture the stylistic transition trend from the raw reference toward the target video. This provides the video stream with high-quality, scene-adapted conditioning signals.
However, integrating these diverse conditioning signals often leads to feature entanglement with the target latents. This entanglement causes style leakage, which severely degrades the unedited regions of the target video. To resolve this, we propose a novel Dual-World-View RoPE (Dual-RoPE). Existing methods typically process condition embeddings by adding task-indicator dimensions~\cite{vace,omniinsert} or treating them as additional frames~\cite{fulldit}, both of which incur substantial training overhead. Dual-RoPE bypasses this limitation by applying distinct coordinate offsets. Specifically, target latents are anchored at zero-offset positions, whereas strong and weak conditions are offset spatially and temporally, respectively. This elegantly isolates all features, allowing the model to effectively process each component without heavy computational burdens.

To address the scarcity of paired data capturing both raw and harmonized object states, we develop a comprehensive data curation pipeline coupled with a rigorous dual-verification mechanism. Specifically, our pipeline automatically synthesizes the requisite training quadruplets from massive open-source text-to-video datasets. Each quadruplet consists of a source video, a target video, an unprocessed reference image, and a harmonized reference image, accompanied by tailored instruction prompts and detailed video captions. Here, the unprocessed reference image serves as the raw object exhibiting severe domain gaps, while the harmonized reference represents the ideal object seamlessly integrated into the target environment.
Leveraging this pipeline, we construct a large-scale, high-quality dataset that has undergone strict evaluation via our automatic dual-verification strategy. Our project page is \href{https://github.com/massyzs/smart-insertion-v}{\textbf{here}}.

In summary, the main contributions of our work are three-fold:

\begin{itemize}

\item \textbf{Decoupled Guidance Module (DGM):} We introduce DGM to equip the framework with comprehensive spatial perception and style discrepancy understanding. It integrates a Vision-Language Model (VLM) for spatial-style reasoning, while independently retaining the native T5 encoder to preserve temporal dynamics and motion priors.

\item \textbf{Dual-Stream Framework:} We propose a \textbf{dual-stream architecture} with \textbf{closed-loop feedback} to concurrently perform video insertion and style transfer for adaptive refinement. We further introduce \textbf{Scope-Constrained Semi-Attention} to prevent undesired condition interactions, and \textbf{Dual-RoPE} to disentangle diverse conditioning signals into strong and weak conditions, achieving high-fidelity results without heavy training overhead.

\item \textbf{Scalable Data Curation and High-Quality Dataset:} We develop an automated, highly scalable data curation pipeline capable of leveraging massive text-to-video datasets. Based on it, we construct a high-quality  dataset.
    
\end{itemize}

\section{Related Work}
\label{sec:related}

\subsection{Video Generation}

Driven by recent algorithmic breakthroughs, the landscape of video generation has undergone a profound paradigm shift. While proprietary systems like SeedDance~\cite{seedance} and Pika~\cite{pika} initially set the benchmark for visual fidelity and spatiotemporal consistency, the recent surge of high-performance open-source foundations—such as HunyuanVideo~\cite{hunyuanvideo}, CogVideo~\cite{cogvideox}, and the Wan series—has rapidly democratized high-fidelity video synthesis, bridging the performance gap with state-of-the-art commercial alternatives. 

\begin{figure*}[th!]
    \centering
    \includegraphics[width=0.92\textwidth,height=5.4cm]{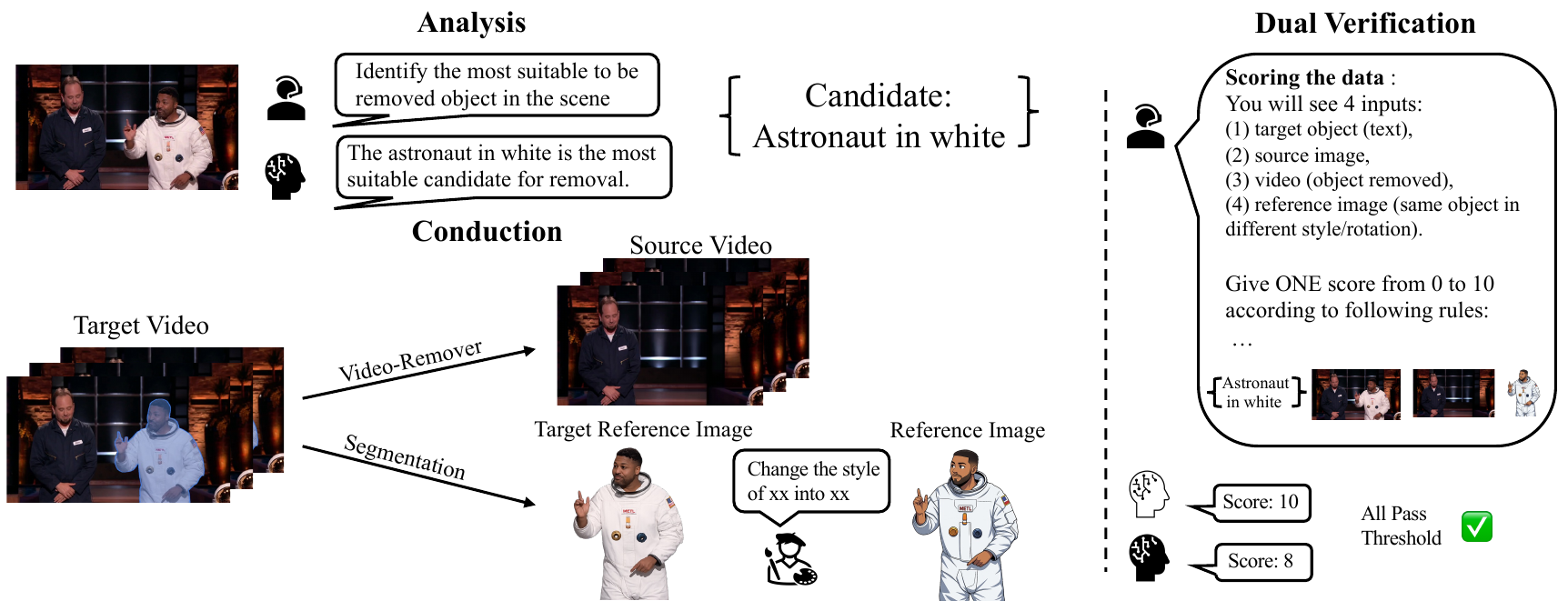}
    
\caption{\textbf{Overview of the data curation pipeline.} We first identify the optimal object within a video that is most suitable for removal. Segmentation and video object removal techniques are then utilized to obtain the source video and the GT reference images. Next, we apply style transfer to the GT references to synthesize the raw, unharmonized references. Finally, all generated data undergo a strict filtering process driven by an automated dual-agent verification.}
    \label{fig:data_curation}
\end{figure*}

\subsection{Video Insertion}

Image-based video insertion aims to seamlessly integrate and animate a reference subject within a source video, necessitating precise spatial localization, robust identity preservation, and genuine contextual harmonization. Existing methods predominantly rely on explicit spatial priors to achieve this. For instance, GetInVideo~\cite{getinvideo} demands impractical, shape-exact masks that severely hinder natural stylistic blending. To relax such stringent spatial constraints, VideoAnyDoor~\cite{videoanydoor} utilizes coarse masks coupled with point trajectories. However, this explicit trajectory conditioning frequently disrupts the intrinsic motion priors of diffusion models, leading to spatiotemporal incoherence where the inserted subject moves asynchronously or disproportionately slower than native scene elements. Exploring alternative paradigms, Anything-in-Any-Scene~\cite{anything} embeds 3D meshes into 3D-lifted video environments. Yet, it still mandates explicit and precise insertion coordinates and suffers from profound material and textural domain gaps compared to the native scene, despite simulating physical illumination. Recently, OmniInsert~\cite{omniinsert} introduced a mask-free approach driven solely by text instructions to bypass these spatial priors. Nevertheless, it fundamentally lacks joint multimodal comprehension; by failing to holistically analyze the video context to modulate the reference subject prior to insertion, it produces rigid, contextually isolated generations. To surmount these limitations, our Smart-Insertion-V framework leverages advanced vision-language reasoning to proactively adapt the subject's appearance and autonomously infer physically plausible insertion layouts without relying on explicit spatial guidance.

\section{Data Curation}
\label{sec:data_curation}

Training the proposed Smart-Insertion-V framework necessitates a two-stage optimization paradigm: (1) pre-training the Vision-Language Model (VLM) adapter on large-scale text-to-video datasets to align multimodal feature spaces, and (2) fine-tuning the entire architecture for photorealistic insertion. To overcome the severe scarcity of paired training data for the fine-tuning phase, we construct a high-fidelity benchmark comprising rigorously aligned data quadruplets (i.e., reference image, target image, source video, and target video). These quadruplets are meticulously synthesized from two distinct sources and subsequently filtered by our dual-verification, which we introduced in supplementary material. The created dataset, based on the public dataset, contains around 0.7M data quadruplets and will be released.

\subsection{Adaptation of Existing Video Editing Datasets}
We repurpose subsets from existing video insertion and removal datasets~\cite{openve}, which typically consist of reference images, input videos, and corresponding ground-truth (GT) videos.
We generate quadruplets through three core operations:
\begin{itemize}
    \item \textbf{Subject Grounding:} We employ Gemini-3-Pro~\cite{gemini3} to describe the object in reference images.
    \item \textbf{Text-Driven Spatiotemporal Segmentation:} Guided by these descriptions, LangSAM~\cite{langsam} is leveraged to extract accurate object masks—from the input videos for insertion tasks, and from the GT videos for removal tasks—ensuring structural integrity.
    \item \textbf{Occlusion-Aware Appearance Recovery:} Since objects extracted from dynamic videos frequently suffer from severe occlusion or frame truncation, we utilize the Nano-Banana~\cite{nanobanana} to proactively hallucinate and reconstruct missing regions. 
\end{itemize}

\subsection{Scalable Generation from Text-to-Video Corpora}
To infinitely scale our dataset and explicitly force the model to learn visual harmonization, we pioneer a secondary curation pipeline built upon unannotated raw video collections as shown in Figure~\ref{fig:data_curation}. We first deploy a VLM to autonomously identify optimal candidate objects for removal. To guarantee high-fidelity downstream processing, candidates must satisfy strict geometric and temporal priors~\cite{wu2024towards,cao2022multi,xin_cyclegrpo2026}: they must not spatially dominate the frame and must exhibit persistent temporal visibility.

Once the initial mask is extracted via LangSAM~\cite{langsam}, we feed it into Minimax-remover~\cite{minimax_remover}, a robust video propagation and inpainting model to remove the subject, yielding a pristine source video. Crucially, to construct a substantial stylistic domain gap between the reference and the target space, we apply a randomized, template-driven style transfer to obtain raw references. Nano-Banana~\cite{nanobanana} systematically alters the material properties and artistic rendering of the subject while strictly preserving its underlying structural conditioning and core identity. Following the completion of occluded regions~\cite{tan2024occluded,chen2023class}, Gemini 3 Pro generates spatial-aware insertion prompts. This deliberate injection of stylistic discrepancy prevents the network from learning trivial copy-paste shortcuts.

\begin{figure*}[t]
    \centering
    \includegraphics[width=\textwidth]{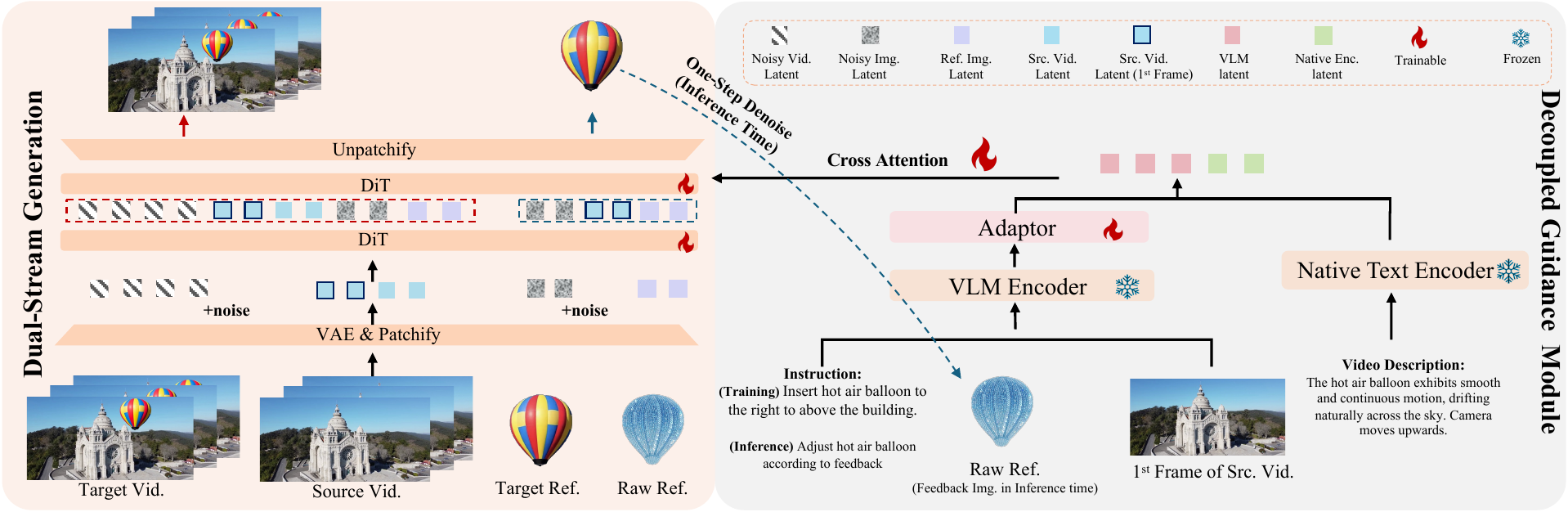}
\caption{\textbf{Overview of the Smart-Insertion-V pipeline (fine-tuning stage).} The framework comprises an image style transfer stream and a video insertion stream. The Decoupled Guidance Module (DGM) processes the raw reference and the first frame of the source video to generate guidance embeddings. Both streams then operate simultaneously: while the video stream conducts the insertion task, the image stream synthesizes harmonized reference features based on $1^{st}$ frame of source video. These image-stream intermediate features are injected into the video stream, providing crucial stylistic guidance to bridge the domain gap. Note that the shared latents between the two streams are drawn separately for illustration purposes only. During inference, the DGM leverages predictions from the image stream as feedback signals to refine the video generation path.}
    \label{fig:pipeline}
\end{figure*}

\section{Method}
\paragraph{Task Formulation.}
We formally define the task of style-adaptive, mask-free video object insertion. Given an unedited source video $V_{src}$, an unprocessed reference image $I_{ref}$, a base video description $P_{desc}$, and a coarse insertion instruction $P_{insert}$, our objective is to synthesize a high-fidelity target video $\hat{V}_{tar}$, yielding a style-harmonized reference image $I_{tar}$ (i.e., target reference image mentioned earlier) as a byproduct. Specifically, the generated $\hat{V}_{tar}$ is expected to integrate the reference object into the target video, satisfying two primary criteria: (1) consistent stylistic harmonization with the source environment, and (2) semantically plausible spatial placement as dictated by the contextual semantics and $P_{insert}$. 


\paragraph{Overview}
Smart-Insertion-V consists of three major components: (1) a Closed-Loop Feedback Dual-Stream framework (Section~\ref{sec:training}), (2) a Decoupled Guidance Module (Section~\ref{sec:guidance}), and (3) Dual-World-View RoPE (Dual-RoPE) (Section~\ref{sec:rope}). The framework jointly performs video insertion and image style transfer during training. During inference, the framework incorporates a feedback signal derived from intermediate predictions to refine video generation path. The Decoupled Guidance Module consists of a native text branch and a VLM reasoning branch, serving as the core guidance of the framework. Dual-RoPE addresses the entanglement issue among target latents and conditions, enabling efficient training and superior performance. We further modify the attention layers of the DiT backbone with a scope-constrained semi-attention mechanism to prevent unintended information leakage across different latent streams.

\subsection{Dual-Stream Framework with Feedback}
\label{sec:training}

Smart-Insertion-V is a dual-stream framework that jointly optimizes a video insertion task and an image style transfer task. The image style transfer stream aims to synthesize style-aligned conditions for the video stream based on the first frame of the source video, while also serving as a feedback signal for the closed-loop mechanism during inference. The generation process of this stream can be formulated as:
\[
\hat{I}_{tar} = DiT(V_{src}^0, I_{ref}, S_{style}),
\]
where $DiT$ refers to our backbone, $V_{src}^0$ denotes the first frame of the source video, and $S_{style}$ denotes the system prompt for adapting the reference image according to the first frame of the source video. The generated $\hat{I}_{tar}$ captures the transformation from the inharmonious reference to the style of the source video, thereby providing effective conditions for the video stream.

The video stream aims to insert the raw reference into the source video. Conditioned on the source video, the raw reference, and the estimated reference features generated by the image stream, this process can be formulated as:
\[
\hat{V}_{tar} = DiT(V_{src}, \hat{I}_{tar}, I_{ref}, P_{insert}, P_{desc}),
\]
where $P_{insert}$ and $P_{desc}$ denote the insertion prompt and the overall video description, respectively. Together with the prompt guidance features from the Decoupled Guidance Module (Section~\ref{sec:guidance}), the reference image and the source video enable harmonized object insertion.

Building upon this dual-task framework, we further propose a closed-loop feedback mechanism for inference time. Specifically, at a denoising timestep $t$, we obtain a one-step estimate $\hat{x}_0^{\text{onestep}}$ from the image-stream alongside the standard image denoising process. Since the video-stream is explicitly conditioned on the intermediate features of the image stream, this estimated harmonized image serves as a reliable indicator of the current insertion quality. We then feed $\hat{x}_0^{\text{onestep}}$ into the VLM branch in the DGM to identify potential stylistic or spatial imperfections. The VLM then produces refinement guidance embeddings to replace the initial generated ones, which guide both streams to modify accordingly. Each feedback loop introduces only approximately 8.14 seconds of additional computation, accounting for merely 3\% of the total inference cost. More on feedback mechanism can be found in supplementary material.


\subsection{Decoupled Guidance Module}
\label{sec:guidance}

Our guidance module consists of two components: (1) a VLM reasoning branch and (2) a native text branch (i.e., T5). Within our framework, these two branches play distinct yet complementary roles. The VLM reasoning module is primarily tasked with high-level comprehension. Specifically, for the image style transfer task, it infers how to adjust the style of the reference image according to source video. For the video insertion task, it interprets coarse insertion instructions to determine a suitable insertion location and adjust object attributes such as scale and orientation. In contrast, the native T5 branch processes general video captions and focuses on motion guidance.

\begin{figure*}[h]
    \centering
    \includegraphics[width=0.9\textwidth]{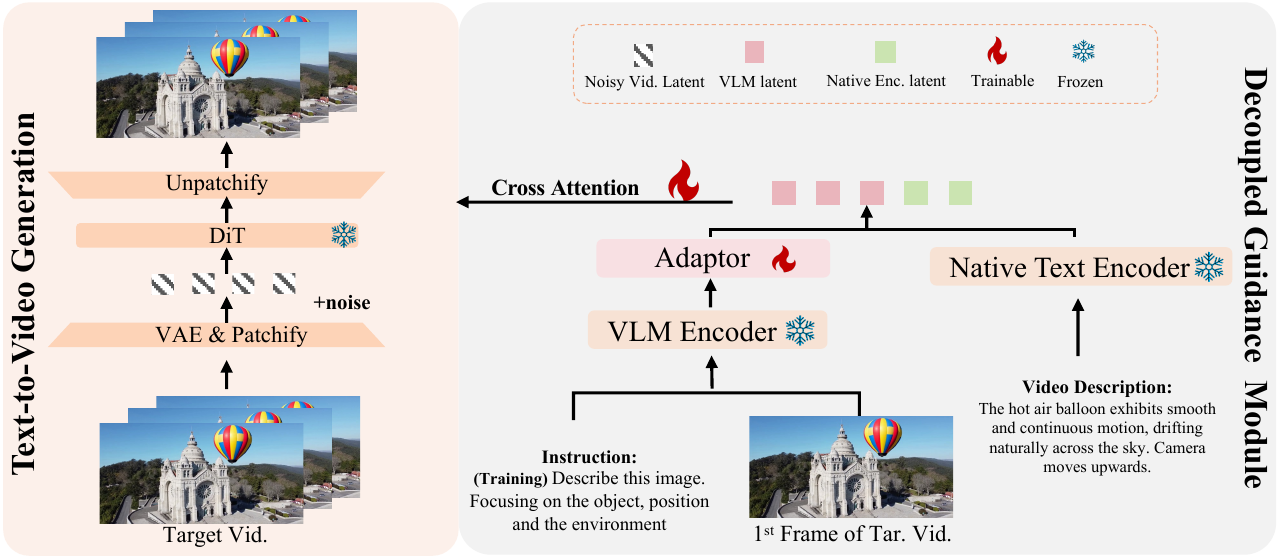}
\caption{\textbf{Decoupled Module Pretraining Stage.} This stage aligns the VLM feature space with the video generation space via image captioning tasks, optimized on text-to-video datasets. Specifically, the VLM processes the first frame of the source video to generate descriptions. This process trains an adapter to effectively map VLM image tokens into the generation space. Finally, the video branch utilizes both the text and the adapted VLM embeddings as guidance to synthesize the corresponding videos.}
    \label{fig:vlm_pretrain}
\end{figure*}

To align the feature spaces of the VLM and the video generation backbone, we introduce an MLP-based adapter. Since directly aligning the latent representations of two large-scale models is highly data-intensive, we strategically decouple the alignment process into two phases: (1) text-to-video pretraining stage (Figure~\ref{fig:vlm_pretrain}), and (2) finetuning stage (Figure~\ref{fig:pipeline}).

During the pre-training stage, the VLM is forced to describe the first video frame (i.e., environment, lighting and atmosphere). The reasoning features extracted from the VLM are formulated as:
\begin{equation}
    e^{\text{vlm}}_{\text{pretrain}} = \text{VLM}(P^{\text{vlm}}_{\text{pretrain}}, V_{\text{src}}^{0}),
\end{equation}
where $e^{\text{vlm}}_{\text{pretrain}}$ denotes the VLM feature embedding, $V_{\text{src}}^{0}$ denotes the first frame of the source video as the visual condition, and $P^{\text{vlm}}_{\text{pretrain}}$ denotes the pre-training prompt for the VLM.
%
%
We discard the token embeddings corresponding to $P^{\text{vlm}}_{\text{pretrain}}$, and retain only the visual and reasoning features, as the prompt embeddings only contain information specific to the VLM module and provide no benefit to the generation task.
The remaining visual tokens act as additional conditioning signals for the generation branch, while the reasoning features serve as guidance. 
The embedding used in our framework is formulated as:
\begin{equation}
    \epsilon^{\text{vlm}}_{\text{pretrain}} = [e^{\text{vlm}}_{\text{pretrain}\mid i} \mid e^{\text{vlm}}_{\text{pretrain}\mid i} \neq \text{vlm}(P^{\text{vlm}}_{\text{pretrain}})],
    \label{eq:select}
\end{equation}
where $i$ denotes the token index.

During the subsequent fine-tuning stage, we tailor the feature extraction for each stream in framework. For the image style transfer task, the guidance features are extracted by:
\begin{equation}
    e^{\text{style}}_{\text{vlm}} = \text{VLM}(P_{\text{style}}, I_{\text{ref}}, V_{\text{src}}^{0}),
\end{equation}
where $I_{\text{ref}}$ denotes the reference image with a distinct style.
The prompt $P_{\text{style}}$ instructs the VLM to analyze the style differences between the first frame of the source video $V_{\text{src}}^{0}$ and the reference image $I_{\text{ref}}$, and to infer how to adapt $I_{\text{ref}}$.
For the video insertion task, the guidance features are derived via:
\begin{equation}
    e^{\text{insert}}_{\text{vlm}} = \text{VLM}(P_{\text{insert}}, I_{\text{ref}}, V_{\text{src}}^{0}),
\end{equation}
where $P_{\text{insert}}$ denotes the coarse insertion instruction containing a brief location description.
The VLM is required to identify a precise insertion location and infer how to refine object attributes such as scale.
Similar to Equation~\ref{eq:select}, we remove the token embeddings corresponding to $P_{\text{style}}$ and $P_{\text{insert}}$ to obtain the clean embeddings $\epsilon^{\text{vlm}}_{\text{style}}$ and $\epsilon^{\text{vlm}}_{\text{insert}}$. For the feedback guidance embedding, we attach in supplementary material.

Regarding the T5 motion guidance, the native text branch focuses on temporal motion descriptions and produces the motion embedding $\epsilon_{\text{motion}}$. By combining $\epsilon_{\text{motion}}$ and $\epsilon^{\text{vlm}}_{\text{insert}}$, we obtain the final guidance embedding for the video insertion stream. 



\begin{figure*}[!h]
    \centering
    \includegraphics[width=0.95\textwidth,height=5.4cm]{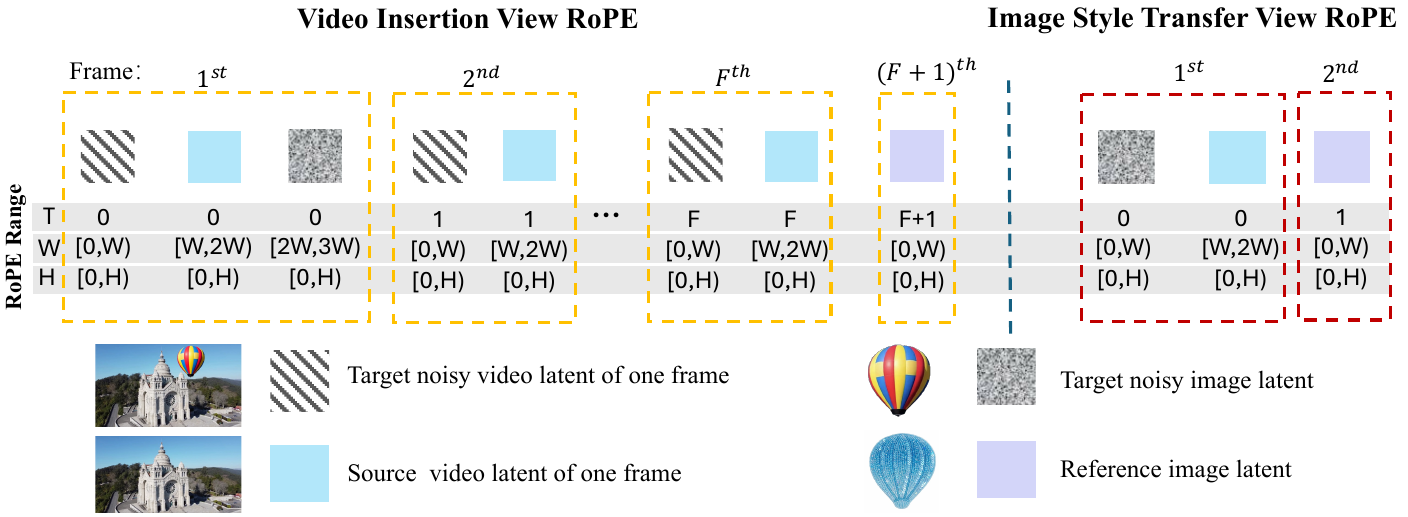}
    \caption{\textbf{Illustration of Dual-World-View RoPE.} The target latents are anchored at zero-offset positions. Strong conditions (e.g., target video and reference image latents) are offset spatially, whereas weak conditions are offset temporally. Notably, this Dual-RoPE strategy is task-agnostic and can be seamlessly applied to both processing streams without modification.}
    \label{fig:dual_rope}
\end{figure*}

\subsection{Dual-world-View RoPE Position Embedding}
\label{sec:rope}

In conditional video generation and editing, the entanglement among noisy target latents and various conditions remains a critical challenge. This issue is particularly pronounced within our proposed dual-task framework. Dual-World-View Rotary Position Embedding, denoted as Dual-RoPE, is proposed to effectively segregate the noisy target latent from the conditioning latents by employing disjoint RoPE operating ranges. Specifically, within the Dual-RoPE framework, we assign a zero-offset strategy to the noisy target latent, while applying tailored offsets along either spatial (width) or temporal dimensions for different conditioning signals. Dual-RoPE allocates task-specific ranges (e.g., for target latents) and maps identical conditions to distinct spaces across the two streams, forming a "Dual-World-View."

In the context of video editing task, we assign distinct spatial-temporal offsets to differentiate the inputs: $\text{Offset}_{\text{src}} = (F=0, W=w, H=0)$ for the source video, $\text{Offset}_{\text{tar}}^{v} = (F=0, W=0, H=0)$ for the target noisy video latent, and $\text{Offset}_{\text{tar}}^{i} = (F=0, W=2w, H=0)$ for the target reference image (from image-stream). Here, $F$, $W$, and $H$ correspond to the temporal, width, and height axes, respectively, whereas $w$ and $N$ denote the spatial width of the inputs and the number of video frames, respectively. The rationale behind this offset design is deeply rooted in the physical intuition of the inputs. First, given the maximum structural similarity between the source and target videos, their positional embeddings are kept in close proximity. Second, spatial offsets provide stronger conditioning signals~\cite{omnitransfer} compared to temporal ones. Consequently, we employ the spatial offset along the $W$-axis to supply the target latent with fine-grained stylistic information of the inserted objects. In contrast, applying a temporal offset to the reference image ensures it acts as a weak condition, offering coarse structural guidance during the initial denoising steps.

\begin{table}[h!]

	\centering
	\small 
	\setlength{\tabcolsep}{3.8pt} 
    \caption{Quantitative results. Ours outperforms all baselines across all metrics. * indicates the cascaded strategy.}
   
    \resizebox{\columnwidth}{!}{%
	\begin{tabular}{l|c c c c|| c c c c|c|c}
		\hline
		\hline
		Metrics\textbackslash Model & UniVideo & UniVideo-Q & TeleOmni &VACE&  UniVideo* & UniVideo-Q* & TeleOmni*  & VACE* & Runway& Ours \\
		\hline
		FVD(/100) $\downarrow$ & 1.687&	1.389&	2.202& 1.302  &	1.888&	1.721&1.807& 1.189 &1.697&\textbf{1.055}\\[1pt]
		Video-LPIPS $\downarrow$ &0.283&	0.246&	0.373& 0.235	& 0.286& 0.307 & 0.358& 0.232 &0.218&\textbf{0.125}\\[1pt]
		DINO-Similarity-V $\uparrow$ &0.629&	0.709&	0.564& 0.738 &	0.594&	0.635 & 0.628 & 0.739&0.727 &\textbf{0.746} \\[1pt]
		Harmonic Score$\uparrow$& 0.661 &0.693 & 0.616& 0.526 & 0.582&0.634 &0.274 & 0.560&0.540&\textbf{0.842}\\[1pt]
       Human(Pos\&Scale)$\uparrow$
& 3.130 & 2.970 & 2.330 & 2.400 & 2.800 & 3.000 & 3.000 & 2.330 & 3.430 & \textbf{4.200}\\[1pt]

Human(Style)$\uparrow$
& 2.500 & 2.970 & 2.330 & 3.330 & 2.870 & 2.970 & 2.800 & 3.000 & 3.000 & \textbf{4.500}\\[1pt]

Human(Motion)$\uparrow$
& 2.700 & 3.170 & 2.030 & 2.470 & 3.000 & 3.430 & 3.300 & 2.700 & 3.470 & \textbf{4.100}\\[1pt]

Human(Avg)$\uparrow$
& 2.777 & 3.037 & 2.230 & 2.733 & 2.890 & 3.133 & 3.033 & 2.677 & 3.300 & \textbf{4.267}\\[1pt]
		\hline
	\end{tabular}}

	\label{tab:results}
\end{table}

\begin{figure*}[!h]
    \centering
    \includegraphics[width=\textwidth]{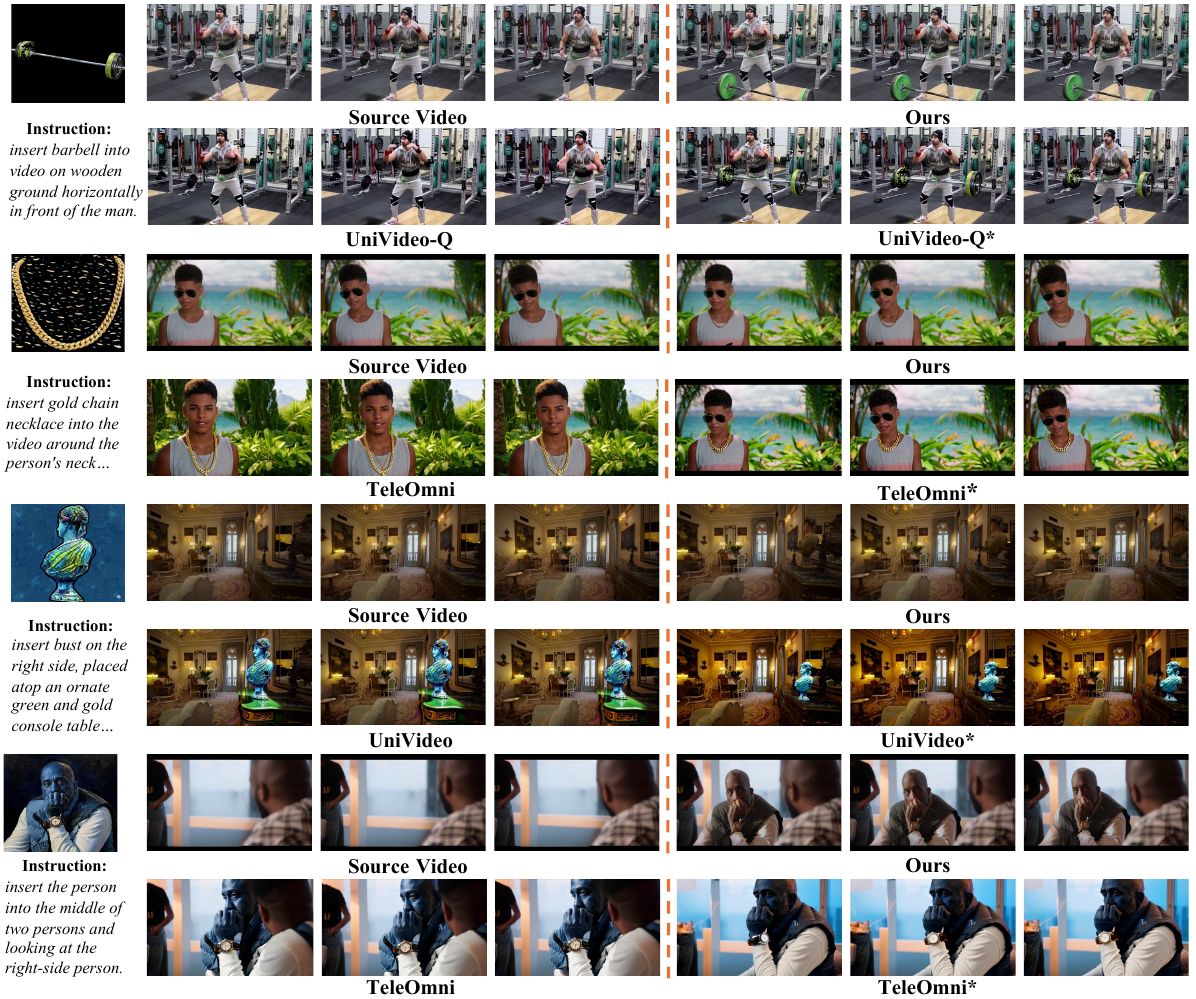}

    \caption{\textbf{Qualitative results}. Our method achieves the best overall performance among six baselines. In contrast, the baselines often fail to harmonize the inserted object with the scene, properly adjust object attributes, or identify a suitable insertion location. We show two representative baselines for each example while ensuring broad coverage of all competing methods across the figure. Full comparisons with all baselines are provided in the supplementary material. The note $*$ refers to baseline cascaded with image editing.}
    \label{fig:results}

\end{figure*}

Following the same design principle, when addressing the image style transfer task, we set the target image to the canonical position without any offset, denoted as $\text{Offset}_{\text{tar}} = (F=0, W=0, H=0)$. To provide precise stylistic guidance, we apply spatial offset along the $W$-axis to the first frame of source video, yielding $\text{Offset}_{\text{src}} = (F=0, W=w, H=0)$. The raw reference is restricted to a temporal shift, formulated as $\text{Offset}_{\text{ref}} = (F=1, W=0, H=0)$. Although there is partial overlap in the RoPE~\cite{rope} spaces between the two streams, the image stream operates independently, thereby preventing any mutual interference.
The Dual-RoPE mechanism is illustrated in Figure~\ref{fig:dual_rope}.

\subsection{Scope-Constrained Semi-Attention}
To prevent unintended information leakage among the target latents and different conditioning signals, we modify the self-attention layers of the DiT backbone with a scope-constrained semi-attention mechanism. Specifically, the target latents are allowed to query the keys and values from all conditioning signals, while each conditioning signal is restricted to attend only within its own scope, without accessing either the target latents or other conditioning signals. This asymmetric attention pattern enables the target latents to aggregate complementary conditioning information while keeping different conditions isolated from one another. In contrast, mainstream full-attention designs~\cite{univideo,fulldit,harminization3} allow unrestricted interactions among all tokens, which may introduce cross-condition information leakage and cause the condition representations to drift toward the target distribution. Our design mitigates such condition-shift and cross-condition interference while preserving effective conditioning. 

\section{Experiment}

In this section, we comprehensively evaluate our proposed Smart-Insertion-V against four representative baselines: TeleOmni~\cite{teleomni}, UniVideo~\cite{univideo} utilizing the last hidden layer feature, UniVideo equipped with a Q-Former~\cite{qformer}, VACE~\cite{vace}, and close-source Runway-Video-Editor (Aleph2.0)~\cite{runway2026aleph2}. Furthermore, to substantiate the inherent limitations of naive cascaded pipelines (i.e., performing explicit image style transfer prior to video object insertion; here, we use Qwen-Image-Edit-2511), we construct augmented cascaded variants for each baseline and discuss them in section~\ref{sec:cascaded}. Detailed implementation settings and evaluation protocols are provided in the supplementary material.

\begin{figure*}[t]
    \centering
    \includegraphics[width=0.98\textwidth]{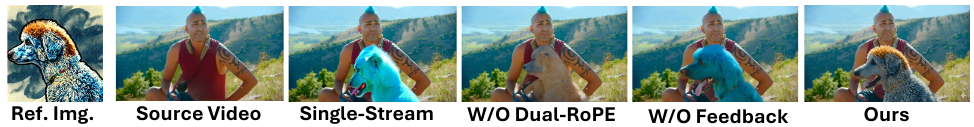}
    \caption{Ablation studies on the effectiveness of dual-stream design, dual-RoPE and closed-loop feedback.}
    \label{fig:ablation}
\end{figure*}
\enlargethispage{\baselineskip}
\subsection{Qualitative Results}
\begin{figure*}[h]
    \centering
    \includegraphics[width=0.94\textwidth,height=21cm]{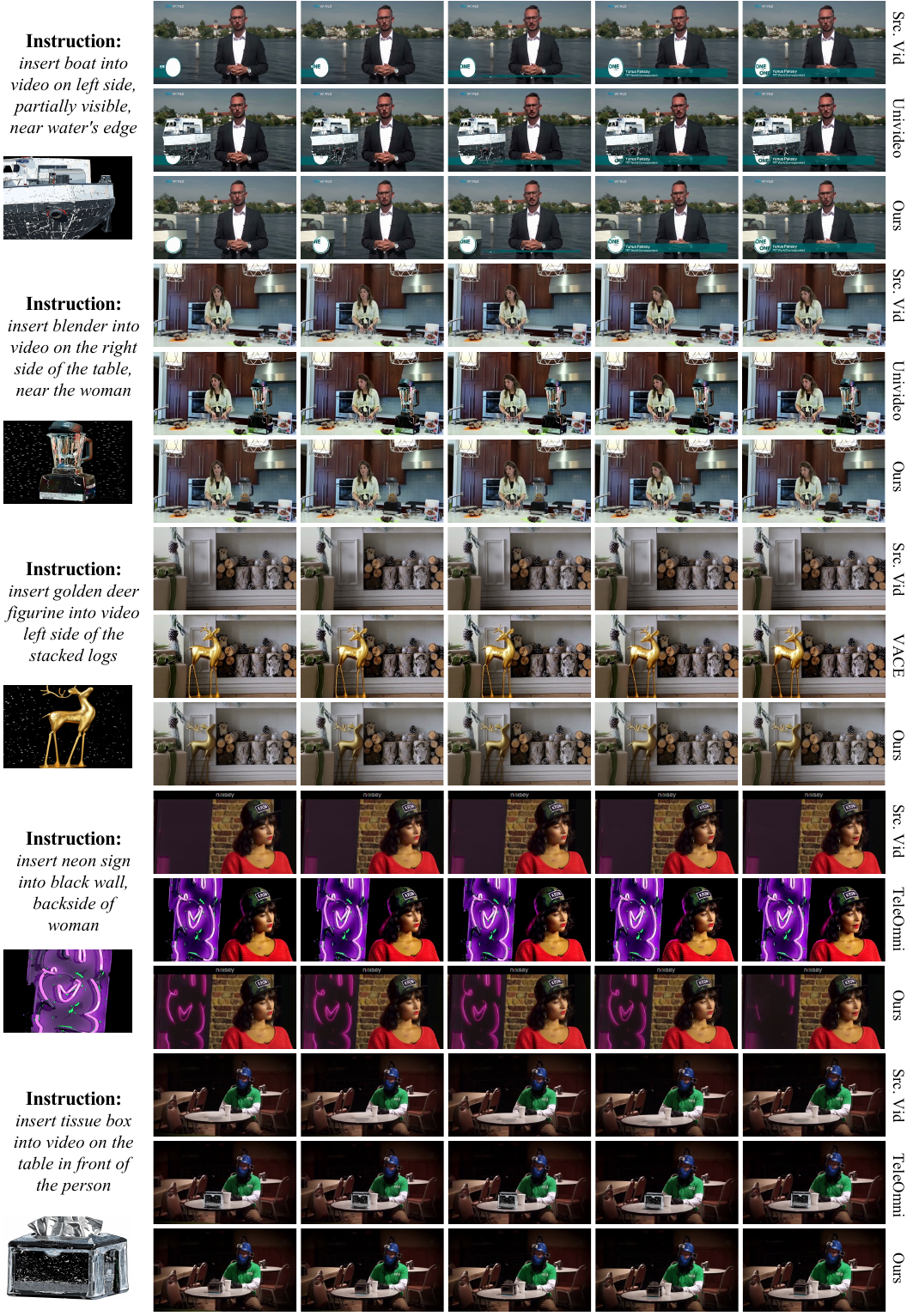}
    \caption{\textbf{More Qualitative Results.} We show more qualitative comparisons with baselines. Ours outperforms them in all cases.}
    \label{fig:more_results}
\end{figure*}

Figure~\ref{fig:results} and Figure~\ref{fig:more_results} present the qualitative results. The visual comparison can be examined from three aspects: (1) the style of the inserted object, (2) the insertion position, and (3) the style of unedited regions. Aspects (1) and (2) reflect the core requirements of the proposed task, while aspect (3) highlights the necessity of Dual-RoPE. Smart-Insertion-V achieves the best qualitative performance in both insertion position and object style harmonization. In contrast, baseline methods often affect the appearance of objects surrounding the inserted object, leading to severe artifacts. Although the cascaded strategy improves baselines, their gains remain limited. More on cascaded strategy can be found in Section~\ref{sec:cascaded}.

\subsection{Quantitative Results}

We evaluate the generated videos using FVD, Video-LPIPS (frame-averaged LPIPS), and DINO-Similarity-V (frame-averaged DINO similarity), using ground-truth videos with harmonized object insertions as references. We also report the harmonic score~\cite{harmonization1,harminization2,harminization3} evaluated by ChatGPT~\cite{chatgpt} and conduct a user study assessing (1) object position and scale, (2) object style, and (3) motion quality, as shown in Table~\ref{tab:results}. Specifically, the harmonic score measures whether the inserted object is stylistically consistent with the source scene, whether it is placed at a reasonable location, and whether its attributes are appropriately adjusted. Detailed evaluation criteria are provided in the Supplementary Material.

\begin{table}[h]

	\centering
	\small 
	\setlength{\tabcolsep}{3.8pt} 
    \caption{Quantitative results of ablation studies.}

    \resizebox{\columnwidth}{!}{%
	\begin{tabular}{l|c c c c}
		\hline
		\hline
		Metrics\textbackslash Ablation & Single-Stream & W/O Dual-RoPE & W/O Feedback & Ours  \\
		\hline
		FVD(/100) $\downarrow$ & 1.364&	1.837&	1.074&	1.055\\[1pt]
		Video-LPIPS $\downarrow$ &0.129&	0.126&	0.141& 0.125	\\[1pt]
		DINO-Similarity-V $\uparrow$ &0.706 &	0.734&	0.749& 0.746	\\[1pt]
        Harmonic Score$\uparrow$&0.474&0.563&0.662&0.842\\[1pt]
Human(Pos\&Scale)$\uparrow$
& 2.900 & 3.300 & 3.700 & 4.200\\[1pt]

Human(Style)$\uparrow$
& 3.100 & 2.970 & 3.700 & 4.500\\[1pt]

Human(Motion)$\uparrow$
& 3.470 & 3.430 & 3.930 & 4.100\\[1pt]

Human(Avg)$\uparrow$
& 3.157 & 3.233 & 3.777 & 4.267\\[1pt]
		\hline
	\end{tabular}}

	\label{tab:ablation}
\end{table}

\enlargethispage{\baselineskip}

\subsection{Discussion on Cascaded Paradigms}
\label{sec:cascaded}
Although sequentially adapting the source image $I_{src}$ to the video $V_{src}$ before insertion seems straightforward, this cascaded approach introduces critical bottlenecks. As demonstrated by applying a powerful image editor prior to baseline insertion methods (Figure~\ref{fig:results} and Table~\ref{tab:results}), such pipelines frequently cause identity distortion and contextual disharmony. We attribute these failures to two fundamental flaws. First, the pipeline suffers from \textit{error accumulation}. Initial editing artifacts and inaccurate style adaptations are inevitably propagated and amplified during subsequent video generation. Second, the decoupled models lack \textit{unified training objectives}. Their independent optimization yields distinct generative priors and feature distributions. This creates a severe domain gap between the intermediate edited image and the insertion model's expected input, ultimately breaking visual coherence. These mechanistic flaws highlight the necessity for our unified, end-to-end framework.

\subsection{Ablation Studies}
In this section, we study the effectiveness of the proposed dual-stream framework, closed-loop feedback inference strategy, and Dual-RoPE. To evaluate the contribution of the auxiliary stream, we remove the image style transfer loss and optimize only the video stream. To assess the closed-loop feedback strategy, we perform inference without the feedback process. For Dual-RoPE, we replace it with the FullDiT~\cite{fulldit} 3D RoPE. As shown in Table~\ref{tab:ablation}, all these variants lead to performance degradation. We also show the qualitative results in Figure~\ref{fig:ablation} and discuss in supplementary material.

\section{Limitations}
While leveraging a VLM enables autonomous spatial grounding from coarse instructions, this semantic-driven paradigm trades off fine-grained spatial controllability. Moreover, our framework is limited by the capability of the adopted 8B VLM, which may fail to recognize objects or understand complex scenes, resulting in inappropriate insertion locations. When the VLM inaccurately interprets the object style or surrounding context, inconsistent features between the conditioning and generation branches may arise, potentially causing object identity drift. Since dynamic instructions are handled separately from the VLM, challenging occlusion, contact, or complex interaction cases may also lead to inaccurate motion understanding. Finally, due to the photorealistic nature of the task, the method could potentially be misused for unauthorized content manipulation, such as watermark removal, raising copyright and provenance concerns. Future work may incorporate watermarking or provenance mechanisms for responsible use.

\enlargethispage{\baselineskip}

\section{Conclusion}
We observe that mask-free video insertion remains fundamentally challenged by stylistically incompatible raw references. To overcome this, we propose Smart-Insertion-V, a dual-stream framework where an image stream conducts reference style transfer while simultaneously guiding the video insertion process. Equipped with DGM, our model further attains precise spatial and stylistic understanding. More broadly, this work suggests a novel generative paradigm that elegantly leverages intermediate predictions as feedback signals. Supported by our newly curated large-scale dataset and pipeline, we achieve harmonious insertion results.

\clearpage
\bibliographystyle{ACM-Reference-Format}
\bibliography{main}

\clearpage

\end{document}